\pdfoutput=1
\documentclass[letterpaper]{article}
\usepackage[preprint]{aaai2027}
\usepackage[hyphens]{url}
\usepackage{graphicx}
\usepackage{natbib}
\usepackage{caption}
\usepackage{amsmath,amssymb}
\usepackage{booktabs}
\usepackage{array}

\newcommand{\Lk}[1]{block~\the\numexpr13-#1\relax}

\title{Object Context Leaves a Halo:\\
Probing How DINOv2 Background Tokens Depend on Nearby Objects}
\author{Zane Kumar\textsuperscript{\rm 1}, Vishal Jain\textsuperscript{\rm 2}}
\affiliations{
\textsuperscript{\rm 1}St Paul's School, London, UK\\
\textsuperscript{\rm 2}Department of Computing, Imperial College London, UK
}

\begin{document}

\maketitle

\begin{abstract}
Dense uses of vision-transformer features treat each patch token as describing
its own patch, but attention lets objects influence nearby background tokens,
so cropping or tiling an image can change a patch's features without changing
its pixels. How much does a background token's representation depend on the
object beside it,
and how does this dependence change through the layers? We introduce crop-context instability, an
interpretability probe that encodes the same pixels within the full image and within a
patch-aligned crop and measures how far each patch embedding moves. On 900
PASCAL VOC images and four frozen DINOv2 backbones, background 4--16\,px from
an object changes 32--67\% more at its peak block than background at least
64\,px away, and the contrast fades or reverses at the final block. This halo
is learned, being absent in untrained networks, and register tokens change
its character: once image position is controlled, standard backbones show almost no
excess near objects, whereas in both register backbones
background next to objects changes 8--20\% more than distant background at the
same location. Register backbones also change less under cropping overall. For pipelines that crop or tile images, register
backbones give the more stable dense features, and their intermediate blocks
are where to test how object context reaches background tokens. Code:
\url{https://github.com/znkumar/halo-crop-context}
\end{abstract}

\section{Introduction}

Dense uses of vision-transformer features treat each patch token as describing
its own patch, yet pipelines that crop or tile images encode the same pixels
with different surrounding context. Later layers mix in information from
elsewhere in the image, and how that exchange is organised is hard to observe
directly. We ask: how does an object affect the
representations of the background around it, and how does that influence
change with depth?

Existing interpretability analyses of self-supervised ViTs describe where tokens attend, which
objects patch affinities recover, and which tokens store global information
(Section~\ref{sec:related}). To our knowledge, none of them measures how
strongly an individual background token depends on the object next to it, or
at which block that dependence is built and then fades or reverses.

Our probe, crop-context instability, compares two views of the same physical patch: the complete image, and a crop whose borders fall on patch boundaries, so every patch in the crop keeps exactly its original pixels. The compared pixels and their scale are fixed; what changes is the visible surrounding context and the token's position in the input grid. The difference between the two hidden states at a given block is a measurable response to that change. We map where in the image the response is largest, how it evolves across blocks, and which attention computations vary with it. Our contributions are:
\begin{itemize}
\renewcommand{\labelitemi}{$\bullet$}
  \item a halo of crop-context instability around objects in four DINOv2
        backbones, peaking at intermediate blocks and fading or reversing at
        the final block;
  \item a comparison across model size and register tokens showing that the
        halo survives an image-position control only in the register
        backbones, where it is concentrated 4--16\,px from objects;
  \item controls for crop position, local appearance, feature norm, mask
        placement and training;
  \item an exploratory link between the halo and changes in object-sourced
        attention updates, which identifies blocks for intervention experiments.
\end{itemize}

\section{Related Work}\label{sec:related}

\paragraph{Object structure in self-supervised ViTs.}
Self-supervised ViTs are known to encode object structure without labels: DINO attention highlights foregrounds \citep{caron2021emerging}, LOST and TokenCut recover objects from patch affinities \citep{simeoni2021localizing,wang2022self}, and DINOv2, trained with self-distillation and iBOT-style patch objectives, yields strong dense descriptors \citep{amir2022deep,oquab2024dinov2,zhou2022ibot}. These works show that object information is present in patch tokens, but not how much a given background token owes to a neighbouring object.

\paragraph{How information moves between tokens.}
Layerwise analyses show ViT attention mixing local and global information differently with depth \citep{raghu2021vision,walmer2023teaching}, ViTs are robust to occlusion and patch removal \citep{naseer2021intriguing}, and high-norm tokens can store global information that registers absorb \citep{darcet2024registers}. Our probe complements these by measuring the dependence of individual background tokens on visible object context, block by block.

\section{A Probe for Context Sensitivity}\label{sec:probe}

\subsection{Keeping local content fixed}
We use four pretrained DINOv2 backbones, each with 12 blocks: ViT-S/14 and ViT-B/14, with and without register tokens \citep{darcet2024registers,oquab2024dinov2}. ViT-S/14 without registers is the reference model for the controls in Section~\ref{sec:controls}. Images and segmentation masks come from PASCAL VOC 2012 \citep{everingham2010pascal}. Each image is resized to $448\times448$ (a $32\times32$ patch grid); masks are resized with nearest-neighbour interpolation. We sample $280\times280$ crops with top-left corner
\begin{equation}
(a_x,a_y)\in\{0,14,\dots,168\}^2,
\end{equation}
so each crop is a $20\times20$ subgrid and corresponding patches have identical pixels. Anchors are drawn uniformly at random.

For patch $p$ and feature layer $\ell$, let $f_\ell(p\mid I)$ and $f_\ell(p\mid C)$ be its embeddings in the full image and the crop. Crop-context instability is the cosine distance
\begin{equation}
s_\ell(p)=1-\frac{f_\ell(p\mid I)^{\top}f_\ell(p\mid C)}{\|f_\ell(p\mid I)\|_2\,\|f_\ell(p\mid C)\|_2}.
\end{equation}
We take patch tokens after each of the 12 blocks, with the final layer norm applied, and refer to layers by block number; block 12 is the model's output.

The crop changes two things at once: the visible context, and the token's grid position, including the interpolated positional embedding for a $20\times20$ rather than $32\times32$ grid. $s_\ell$ alone cannot separate them. Our claims therefore rest on \emph{contrasts} between near-object and distant tokens, which we also evaluate at identical crop-grid coordinates and at identical image-grid coordinates (Section~\ref{sec:controls}, Table~\ref{tab:family}).

\subsection{An object-relative coordinate system}
Masks are used only to define evaluation groups. A patch is background if most of its pixels are background, and valid if most lie outside void regions. Each background patch is assigned the mean pixelwise Euclidean distance to the nearest foreground pixel. We compare patches 4--16\,px from an object (near) with those at least 64\,px away (far).

The main experiment uses 900 validation images and four crops per image: 1{,}400{,}319 valid patch observations, of which 123{,}039 are near and 362{,}752 far. We report group means, their ratio, and the AUROC for ranking near above far. Because patches and crops from the same image are dependent, all intervals and tests are computed at the image level.

\section{The Halo and Its Depth Profile}\label{sec:halo}

\subsection{An object-aligned band of context dependence}
\begin{figure}[t]
\centering
\includegraphics[width=\columnwidth]{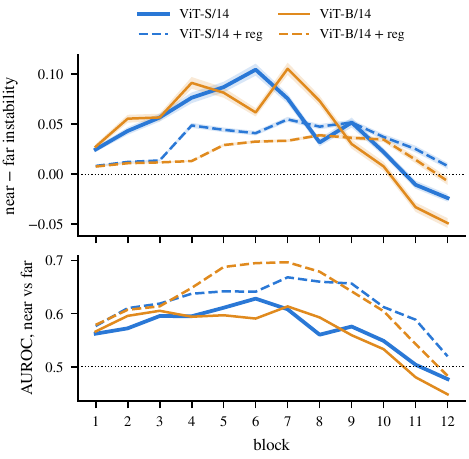}
\caption{Depth profile of the near--far contrast in four DINOv2 backbones (900 images). Top: mean per-image difference in instability between near and far background, with 95\% image-level bootstrap bands. Bottom: AUROC for ranking near above far background tokens. Block 12 is the model output.}
\label{fig:depth}
\end{figure}

\begin{figure*}[t]
\centering
\includegraphics[width=\textwidth]{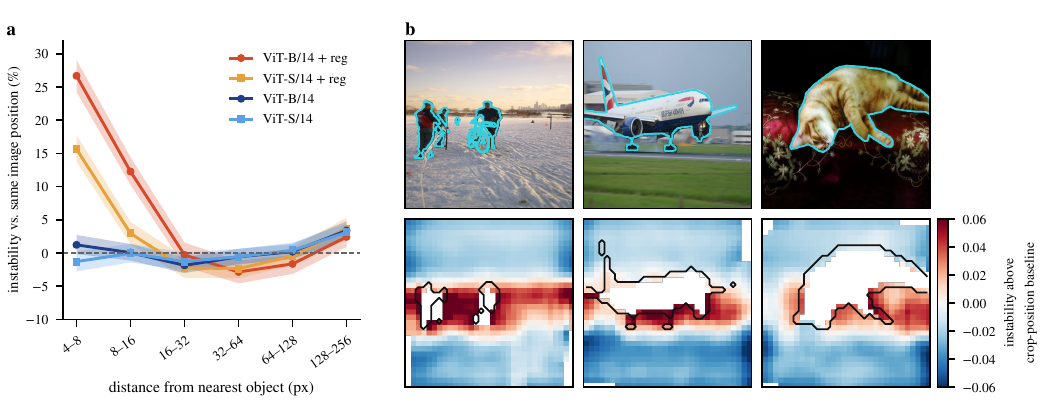}
\caption{\textbf{Background next to objects is most sensitive to context.} (a) Instability at block 5 relative to the dataset mean at the same image position, by distance from the nearest object, as a percentage of the model's mean instability. Lines are the four backbones: means over the 900 images of the main run with 95\% image-level bootstrap bands. (b) Three VOC images with objects outlined, and below each its instability at block 5 in ViT-B/14 with registers, after removing the crop-position term. Maps average 24 patch-aligned crops and are $3\times3$-smoothed for display; white tokens are excluded (object, void border or uncovered). The three images were chosen from the 16 with the largest near--far difference among 140 candidates, to show different object sizes and positions.}
\label{fig:halo}
\end{figure*}

At intermediate depth, instability is higher in the background around objects in all four backbones (Figures~\ref{fig:depth} and~\ref{fig:halo}). In ViT-S/14 at block 5, mean instability is 0.284 for near and 0.221 for far background (ratio 1.29, AUROC 0.611). The contrast peaks at block 6 (0.310 versus 0.228, ratio 1.36, AUROC 0.628). ViT-B/14 follows the same profile (peak ratio 1.32 at block 7). The register backbones change less overall but have larger near--far ratios: up to 1.57 in ViT-S/14 and 1.67 in ViT-B/14, with AUROC up to 0.668 and 0.696.

\subsection{The contrast at the final blocks}
The near-object effect is not carried to the output. In ViT-S/14 the ratio falls to 0.97 at block 11 (AUROC 0.504) and 0.92 at block 12, so in the output distant background changes more. ViT-B/14 shows a stronger reversal (0.84 at block 12). Across the 867 images with both near and far tokens, the mean per-image difference in ViT-S/14 is $+0.087$ at block 5 (95\% bootstrap interval $[0.082,0.092]$; two-sided sign-flip $p<10^{-4}$), $-0.011$ at block 11 ($[-0.015,-0.006]$) and $-0.024$ at block 12 ($[-0.027,-0.020]$). In the register backbones the contrast also shrinks toward the output but does not clearly reverse: the block-12 ratio is 1.18 in ViT-S/14 and 1.06 in ViT-B/14, where the image-level difference is slightly negative ($-0.007$, $[-0.011,-0.003]$).

\section{Controls on the Spatial Interpretation}\label{sec:controls}

The first three controls use ViT-S/14, each as a separate experiment with its own sample; the last compares the four backbones on the main run (Table~\ref{tab:family}).

\subsection{Position and local appearance}
A position-only predictor reaches AUROC at most 0.585, and horizontal or vertical position alone at most 0.556. Comparing near and far tokens at identical crop-grid coordinates, where both share the same positional-embedding change, the weighted ratios are 1.39 at \Lk{8} and 1.44 at \Lk{6}, with positive differences at 88.5\% and 94.3\% of 400 eligible coordinates. Local RGB gradient alone gives AUROC 0.548; residualising instability on gradient and quadratic position terms leaves AUROC 0.616 at \Lk{8} and 0.618 at \Lk{6}.

\subsection{Object geometry and training}
Circularly shifting the masks by at least 48\,px breaks the alignment between objects and the instability field. On this run the ratio at \Lk{8} falls from 1.35 with true masks to 0.91 with shifted masks, and at \Lk{6} from 1.35 to 0.90, so the contrast depends on where the masks are placed. Shifting masks also moves them away from the image centre, so this control does not separate object location from image position (Table~\ref{tab:family}). On a matched subset, three randomly initialised ViT-S/14 networks give near--far differences between $-0.009$ and $-0.003$, against $+0.074$ for the pretrained model. Random networks barely change under cropping (mean instability 0.01--0.02), so we conclude only that they show no positive near--far contrast; the contrast is learned.

\subsection{Feature norms}
High-norm tokens can carry global information \citep{darcet2024registers}. On the intermediate-layer contrasts, feature norm alone gives AUROC 0.31--0.49: near-object tokens tend to have lower norm, so norm is inversely related to group membership. Residualising instability on gradient, position and norm in both views still leaves an image-level near--far difference of $+0.075$ ($[0.070,0.080]$; AUROC 0.593).

\subsection{Model size, registers and image position}\label{sec:family}

\begin{table}[t]
\centering
\footnotesize
\setlength{\tabcolsep}{2.5pt}
\begin{tabular}{lcccc}
\toprule
 & S/14 & S/14+reg & B/14 & B/14+reg\\
\midrule
Peak ratio (block) & 1.36 (6) & 1.57 (7) & 1.32 (7) & 1.67 (6)\\
Ratio at block 12 & 0.92 & 1.18 & 0.84 & 1.06\\
\midrule
\multicolumn{5}{l}{\emph{Block 5, near vs far at the same coordinate}}\\
Crop grid: ratio & 1.30 & 1.42 & 1.27 & 1.46\\
Image grid: ratio & 1.00 & 1.08 & 1.01 & 1.20\\
Image grid: \% positive & 49 & 88 & 55 & 99\\
\bottomrule
\end{tabular}
\caption{The four backbones on the main run. Matched ratios compare near and far background at identical coordinates, weighted over coordinates with at least 20 observations of each; ``\% positive'' is the share of the 770 eligible image-grid coordinates where near exceeds far.}
\label{tab:family}
\end{table}

The crop-coordinate control fixes a token's position in the crop but not in the image, and VOC objects tend to sit near the image centre, so near-object background is on average more central than distant background. We therefore also compare near and far tokens at identical coordinates of the $32\times32$ image grid, pooling over images (Table~\ref{tab:family}).

The result depends on register tokens. In ViT-S/14 and ViT-B/14 without registers, the ratio at matched image coordinates is 1.00 and 1.01, and near exceeds far at only 49\% and 55\% of coordinates. After subtracting the dataset mean at each image coordinate, the image-level near--far difference is $-0.002$ ($[-0.006,0.002]$) and $0.000$ ($[-0.003,0.004]$). In these backbones the halo of Section~\ref{sec:halo} cannot be distinguished from a dependence of instability on image position.

With registers, the halo survives. The matched ratio is 1.08 in ViT-S/14 and 1.20 in ViT-B/14, near exceeds far at 88\% and 99\% of image coordinates, and the image-level differences after removing the image-position term are $+0.008$ ($[0.006,0.009]$) and $+0.009$ ($[0.008,0.010]$). The excess sits close to objects: with registers it is 27\% and 12\% of the model's mean instability at 4--8 and 8--16\,px in ViT-B/14, 16\% and 3\% in ViT-S/14, and near zero beyond 16\,px (Fig.~\ref{fig:halo}a). A plausible reading is that the high-norm artefact tokens that registers remove \citep{darcet2024registers} dominate crop instability in the standard backbones and hide the smaller object-related effect; we have not tested this.

\section{Object-Sourced Attention Updates}\label{sec:attn}

\subsection{Definition}
For a background query token $p$ in attention block $b$, let $O$ be the foreground tokens, $a^{(b,h)}_{p,o}$ the attention weight from $p$ to $o\in O$ in head $h$, and $v^{(b,h)}_o$ the value vector. The object-sourced update before the output projection is
\begin{equation}
g_b(p\mid I)=\operatorname*{concat}_h\Big(\sum_{o\in O}a^{(b,h)}_{p,o}\,v^{(b,h)}_o\Big),
\end{equation}
with attention normalised over all tokens. We compute $g_b(p\mid C)$ in the crop analogously, with $O$ the foreground tokens inside the crop, and record $\Delta_b(p)=\|g_b(p\mid I)-g_b(p\mid C)\|_2$, the angle between the two updates, and an alignment statistic between the update change and the feature change.

\subsection{Association with instability}
We use 250 images, two crops each, and up to 120 distance-stratified background tokens per crop (about 56{,}000 observations per configuration). A baseline regression predicts $s_\ell(p)$ from position, object distance, both feature norms and both update norms. We then add the three update-change variables and compare $R^2$ over three feature layers and twelve attention blocks per backbone.

\begin{table}[t]
\centering
\small
\begin{tabular}{lcc}
\toprule
 & DINOv2 & + registers\\
\midrule
Feature layer & \Lk{6} & \Lk{4}\\
Attention block & 5 & 10\\
Baseline $R^2$ & .048 & .148\\
With update changes $R^2$ & .269 & .367\\
$\Delta R^2$ & .222 & .218\\
Within-distance shuffle $\Delta R^2$ & .0002 & .0002\\
\bottomrule
\end{tabular}
\caption{Largest association between update changes and instability, selected from 36 layer--block pairs per backbone.}
\label{tab:attn}
\end{table}

The best configurations add 0.222 and 0.218 to $R^2$ (Table~\ref{tab:attn}). From the lowest to the highest $\Delta_b$ decile, mean instability rises from 0.190 to 0.337 ($1.78\times$) in the original model and from 0.071 to 0.196 ($2.77\times$) with registers. Permuting the update-change variables within ten object-distance bins (50 draws) reduces the increment to about 0.0002 (maximum 0.0006), so the association is not explained by distance alone. With object distance in the baseline, the residual near--far gap is small (0.0081 without registers, 0.0034 with); adding the update variables reduces it to 0.0004 and $-0.0028$.

Three caveats apply: the alignment statistic is computed from the feature change itself, so the increments are an upper bound; the values are maxima over 36 configurations without held-out validation; and the best register-backbone attention block (10) lies downstream of the feature it predicts (\Lk{4}), so it reflects co-variation, not cause. Table~\ref{tab:attn} therefore specifies the next experiment: ablate background-to-object attention edges at the selected block, compare matched background-to-background edges, and measure the change in the later-layer halo.

\section{Discussion and Limitations}

The clearest result is the effect of register tokens, which our matched pairs separate from model size (Fig.~\ref{fig:halo}a). After the image-position control, both standard backbones stay within 4\% of zero, while with registers background 4--8\,px from an object changes 16\% (ViT-S/14) and 27\% (ViT-B/14) more than same-position background. Registers bring out a sharp, local dependence on nearby objects and larger models show more of it; their blocks 5--9 are where to study it.

Crop instability mixes visible-context and positional-embedding change, and the image-position control removes only an average position effect. Removing context also removes object content for some tokens, and patches 4--8\,px from an object can contain some object pixels, so the remaining halo may partly reflect boundary content; object-removal and background-replacement interventions would separate these. The ViT-S/14 controls and Section~\ref{sec:attn} predate the image-position control and were not rerun with it. Results cover two model sizes, one crop size and one centre-biased dataset; other ViTs and object layouts would test generality.

\section{Conclusion}

Changing the context of an identical patch probes how a ViT distributes object information, and mirrors cropping and tiling in deployment.
DINOv2 mixes object context into nearby background and then largely removes it,
and this shows only with register tokens: there, background within about
16\,px of an object changes 8--20\% more than distant background at the same
image position, peaking at intermediate blocks and fading toward the output,
whereas without registers the position-controlled profile is nearly flat. For pipelines that crop or tile
images before extracting dense DINOv2 features, register backbones are the more
stable choice, and their intermediate blocks are the place for causal
tests of how object context reaches background tokens.

\bibliography{references}

\begin{thebibliography}{11}
\providecommand{\natexlab}[1]{#1}

\bibitem[{Amir et~al.(2022)Amir, Gandelsman, Bagon, and Dekel}]{amir2022deep}
Amir, S.; Gandelsman, Y.; Bagon, S.; and Dekel, T. 2022.
\newblock Deep {ViT} Features as Dense Visual Descriptors.
\newblock In \emph{ECCV Workshops}.

\bibitem[{Caron et~al.(2021)Caron, Touvron, Misra, J{\'e}gou, Mairal, Bojanowski, and Joulin}]{caron2021emerging}
Caron, M.; Touvron, H.; Misra, I.; J{\'e}gou, H.; Mairal, J.; Bojanowski, P.; and Joulin, A. 2021.
\newblock Emerging Properties in Self-Supervised Vision Transformers.
\newblock In \emph{ICCV}.

\bibitem[{Darcet et~al.(2024)Darcet, Oquab, Mairal, and Bojanowski}]{darcet2024registers}
Darcet, T.; Oquab, M.; Mairal, J.; and Bojanowski, P. 2024.
\newblock Vision Transformers Need Registers.
\newblock In \emph{ICLR}.

\bibitem[{Everingham et~al.(2010)Everingham, Van~Gool, Williams, Winn, and Zisserman}]{everingham2010pascal}
Everingham, M.; Van~Gool, L.; Williams, C. K.~I.; Winn, J.; and Zisserman, A. 2010.
\newblock The {PASCAL} Visual Object Classes ({VOC}) Challenge.
\newblock \emph{IJCV}.

\bibitem[{Naseer et~al.(2021)Naseer, Ranasinghe, Khan, Hayat, Khan, and Yang}]{naseer2021intriguing}
Naseer, M.; Ranasinghe, K.; Khan, S.; Hayat, M.; Khan, F.~S.; and Yang, M.-H. 2021.
\newblock Intriguing Properties of Vision Transformers.
\newblock In \emph{NeurIPS}.

\bibitem[{Oquab et~al.(2024)Oquab, Darcet, Moutakanni, Vo, Szafraniec, Khalidov, Fernandez, Haziza, Massa, El-Nouby, Assran, Ballas, Galuba, Howes, Huang, Li, Misra, Rabbat, Sharma, Synnaeve, Xu, J{\'e}gou, Mairal, Labatut, Joulin, and Bojanowski}]{oquab2024dinov2}
Oquab, M.; Darcet, T.; Moutakanni, T.; Vo, H.; Szafraniec, M.; Khalidov, V.; Fernandez, P.; Haziza, D.; Massa, F.; El-Nouby, A.; Assran, M.; Ballas, N.; Galuba, W.; Howes, R.; Huang, P.-Y.; Li, S.-W.; Misra, I.; Rabbat, M.; Sharma, V.; Synnaeve, G.; Xu, H.; J{\'e}gou, H.; Mairal, J.; Labatut, P.; Joulin, A.; and Bojanowski, P. 2024.
\newblock {DINOv2}: Learning Robust Visual Features without Supervision.
\newblock \emph{TMLR}.

\bibitem[{Raghu et~al.(2021)Raghu, Unterthiner, Kornblith, Zhang, and Dosovitskiy}]{raghu2021vision}
Raghu, M.; Unterthiner, T.; Kornblith, S.; Zhang, C.; and Dosovitskiy, A. 2021.
\newblock Do Vision Transformers See Like Convolutional Neural Networks?
\newblock In \emph{NeurIPS}.

\bibitem[{Sim{\'e}oni et~al.(2021)Sim{\'e}oni, Puy, Vo, Roburin, Gidaris, Bursuc, P{\'e}rez, Marlet, and Ponce}]{simeoni2021localizing}
Sim{\'e}oni, O.; Puy, G.; Vo, H.~V.; Roburin, S.; Gidaris, S.; Bursuc, A.; P{\'e}rez, P.; Marlet, R.; and Ponce, J. 2021.
\newblock Localizing Objects with Self-Supervised Transformers and No Labels.
\newblock In \emph{BMVC}.

\bibitem[{Walmer et~al.(2023)Walmer, Suri, Gupta, and Shrivastava}]{walmer2023teaching}
Walmer, M.; Suri, S.; Gupta, K.; and Shrivastava, A. 2023.
\newblock Teaching Matters: Investigating the Role of Supervision in Vision Transformers.
\newblock In \emph{CVPR}.

\bibitem[{Wang et~al.(2022)Wang, Shen, Hu, Yuan, Crowley, and Vaufreydaz}]{wang2022self}
Wang, Y.; Shen, X.; Hu, S.~X.; Yuan, Y.; Crowley, J.~L.; and Vaufreydaz, D. 2022.
\newblock Self-Supervised Transformers for Unsupervised Object Discovery Using Normalized Cut.
\newblock In \emph{CVPR}.

\bibitem[{Zhou et~al.(2022)Zhou, Wei, Wang, Shen, Xie, Yuille, and Kong}]{zhou2022ibot}
Zhou, J.; Wei, C.; Wang, H.; Shen, W.; Xie, C.; Yuille, A.; and Kong, T. 2022.
\newblock {iBOT}: Image {BERT} Pre-Training with Online Tokenizer.
\newblock In \emph{ICLR}.

\end{thebibliography}

\end{document}